# Acceptable Misalignment

Maud van Lier[1]*, Ludovica Marinucci[2], Vieri Giuliano Santucci[3], Raja Chatila[1], Stéphane Doncieux[1], Émiland Garrabé[1], Richard Duro[4]

[1] Sorbonne Université, CNRS, Institut des Systèmes Intelligents et de Robotique, ISIR, F-75005 Paris, France
[2] Interdepartmental Centre Research Ethics and Integrity (CID Ethics), National Research Council (CNR), Rome, Italy
[3] Institute of Cognitive Sciences and Technologies (ISTC), National Research Council (CNR), Rome, Italy
[4] Integrated Group for Engineering Research, Centro de Investigación en Tecnologías de la Información y las Comunicaciones (CITIC), Universidade da Coruña, Spain

*Corresponding author:
maud.van_lier@isir.upmc.fr

## Abstract
There is a specific challenge designers face in developing autonomous robots capable of open-ended learning. This challenge is how one can develop such systems in an ethically sound way when there are many *unknowns* during design time about their manner of deployment, what they can and will learn, and how these systems will act on the basis of what they have learned. One response to this challenge is to align, as far as possible, the activities and goals of autonomous robots with human values, raising the question of *how* to make this alignment possible, a question known as the *value alignment problem.* We argue that the value alignment problem faces a number of challenges that show that full alignment is both theoretically and practically impossible, meaning that misalignment incidents will keep occurring. This undermines current alignment approaches, as the attempt to maximize alignment cannot properly deal with the possibility that the risk of misalignment that remains after alignment methods have been implemented might still be unacceptable to society. To overcome this difficulty, we argue that a shift in perspective is required from a focus on the maximization of alignment to *the conditions under which misalignment incidents become acceptable*. In this paper, we explore what this shift in perspective entails. We introduce the concept of *acceptable misalignment*, formulate normative requirements for alignment mechanisms aimed at bringing misalignment incidents to an acceptable level, and we conclude by presenting a new methodological framework that consists of alignment mechanisms that can meet these requirements.

## 1. Introduction

With increasing speed, we are advancing towards a future in which we can summon autonomous robots to help us out with a variety of tasks, whether it is to perform routine household chores, to accomplish novel tasks, or to provide assistance in complex and dynamic social settings. While this possibility has the potential to impact our society in a positive way by making our current societal practices more accessible to a broader range of people, enabling greater independence, and extending access to forms of support that may otherwise be difficult to obtain, this positive outcome is not a guarantee. This is because these more advanced forms of automated support require high levels of autonomous functioning that entail relinquishing control over what these systems can learn, what goals they might pursue, and how they are able to execute the tasks that we set them. Without such control, there is the possibility that these systems might develop behaviors that are not useful or that might even harm us, or that they are used for purposes that they were either not intended for or that they cannot technically execute. The question is therefore how we can ensure that we can develop autonomous robots in an ethically sound way, meaning that their impact on society will be positive rather than negative.

The difficulty in answering this question is that there are many *unknowns*: What is it that these systems might learn? For what purposes will they be used? What kind of environments will they be exposed to and how will their introduction into such environments alter the environments themselves? These are questions that cannot be answered by approaches such as *Value Based Design* (Friedman & Hendry 2026) and *Responsible Innovation* (Stilgoe et al. 2020). This is because such approaches are largely evidence-based — they deal with uncertainties rather than unknowns. Even the use of (simulated) scenarios to anticipate possible futures is risky, since we do not know the probability of these scenarios actually happening. The challenge is therefore how we can develop autonomous systems in an ethically sound manner when we can only generate limited operational experience during design time.

One proposal to deal with such unknowns is to *align* the activities of autonomous systems and the goals that they can pursue with human values. This way, or so the thought goes, the system will act in a way that we approve of even in situations that could not be foreseen at design time, since its decision-process will be aligned with our values. The question, then, is *how* we can make this alignment possible, a question more generally known as the *value alignment problem.* As we argue in this paper, the value alignment problem faces a number of challenges that show that full alignment is both theoretically and practically impossible, meaning that misalignment incidents will keep occurring. This poses a problem for current alignment approaches, as the attempt to maximize alignment cannot properly deal with the possibility that the risk of misalignment that remains after alignment methods have been implemented might still be unacceptable to society. To overcome this difficulty, we argue that a shift in perspective is required from a focus on the maximization of alignment to *the conditions under which misalignment incidents become acceptable*.

In this paper, we work out what this shift in perspective entails, including what we mean by acceptability and what kind of alignment mechanisms would enable us to reduce the occurrence of misalignment incidents to an acceptable level. To make this point, we first present the current approach to alignment in more detail and show why a change in perspective is needed (Section 2). We then go deeper into what we mean by acceptable misalignment and what criteria alignment mechanisms would need to meet to make misalignment incidents acceptable (Section 3). Based on these normative criteria, we introduce our framework, which we call the *Acceptable Misalignment Framework,* as a new approach to develop autonomous systems in an ethically sound way (Section 4). We end the paper by going over possible limitations and by presenting a future outlook on how the framework could be used (Section 5).

## 2. Socially Disruptive Technologies

There are certain technologies of which we can already tell that their impact on society will be profound. This is because their introduction into society will challenge the way that we do things, how we interact with each other, what we can know, etc. What their exact impact will be, however, is hard to determine beforehand, as we cannot simulate society in all its complexity in a laboratory setting. As a result, the full extent of their impact on society remains largely unknown and will only become apparent once they are introduced into it. These technologies are commonly referred to as *socially disruptive technologies* (van de Poel et al. 2023), and the main challenge they pose is how they can be developed in an ethically sound manner when their societal impact is not just uncertain but mostly unknown during the development process.

In this section, we claim that autonomous robots, and then specifically those capable of *open-ended learning,* fall within this domain of socially disruptive technologies and that their development therefore faces the same challenge (Section 2.1). We further argue that one proposed solution to deal with this challenge, to align the activities of these systems as well as the goals that they pursue with society's values, is, in its currently proposed form, unable to meet this challenge, and that therefore a shift in perspective is required (Section 2.2).

### 2.1 Autonomous Robots

Autonomous robots are embodied artificial systems that possess some capability to make and act on independent decisions when interacting with the physical or virtual environment that they are embedded in. Here we understand 'embodiment' in the broad sense of a "*medium* through which [a system] is able to interact with an environment" (van Lier 2026, p. 14, italics are ours). Autonomous robots should thereby be differentiated from *automated* systems, which only execute predetermined sequences of actions, as in the case of a driver-less underground train. While automated systems may respond in different ways to a variety of inputs or contexts, they always do so according to prescriptive rules that admit no deviation. They thus neither possess nor can independently develop goals of their own, and because they lack the capacity to learn they are not capable of intelligent problem solving.

What characterizes autonomy proper is the capacity of independent decision making, which usually leads a system to learn and to modify itself over time, and it is in this sense connected to machine learning. Autonomy is thereby not an all or nothing property, since it can be exercised at different levels of operation. It may be confined to individual policies or extend as far as the setting of final goals.[1] For artificial systems we can distinguish several levels of autonomous operation. At the lower end are systems that can autonomously accomplish assigned tasks, that is, identify the best strategy to achieve an externally set goal. In reinforcement learning (Sutton et al. 1998), for instance, a signal defined by the developer or the user allows the system, which interacts with the world through trial-and-error, to discover the policy that maximizes that signal and thereby accomplishes the task of which the signal is a proxy.

A more advanced form of autonomous operation is achieved by systems that are able to generate sub-goals of their own, through which they set themselves tasks and build competences that can subsequently be combined to solve the assigned task (Jurgenson et al. 2020; Huang et al. 2024). While the final task remains given, its decomposition becomes an autonomous choice of the system, where these choices still rest on predefined heuristics. The two levels can thereby be distinguished by whether the constraints that were fixed at design time determine the content of the behavior of the system or only its form. Prescriptive rules specify what the system will do, whereas sufficiently general heuristics leave undetermined which goals the system will set itself and which competences it will acquire, granting it decisional and not merely behavioral autonomy in environments potentially unknown to developers and users alike.

At the highest level we may place autonomous open-ended learning systems (Open-ended Learning Team et al. 2021; Sigaud et al. 2023), the scope of which is not the solution of a specific task but the life-long acquisition of knowledge and competences concerning the interaction with a dynamic environment. Described differently, the robot is assigned the meta-task of learning as much as possible. One instance of this approach is intrinsically motivated open-ended learning (Santucci et al. 2020), in which signals based on novelty, surprise or competence improvement are autonomously generated to drive the system to discover new goals, in turn serving as targets for the learning of new competences (e.g., Blaes et al. 2019; Forestier et al. 2022; Colas et al. 2022; Fallas-Hernández et al. 2025; Romero et al. 2025). However, such a high level of autonomy is remote from the applied requirements of robotics, where artificial systems are predominantly developed for a specific purpose.

Nonetheless, open-ended learning techniques may prove useful in applied settings. Even when guided by externally assigned tasks, as in intrinsically motivated open-ended learning, they secure the autonomy needed to acquire competences not anticipated at design time, enabling robots to respond to human requirements when tasks are underspecified and/or environments are unknown or dynamic (Morelli et al. 2025). In such settings, the intended purpose bounds the range of contexts of operation, but what the system learns cannot be fully anticipated, meaning that risk identification and safeguard design remain non-trivial.

These difficulties are heightened when open-ended learning mechanisms are implemented in systems designed for general-purpose use. Such autonomous robots may be deployed across diverse domains with different goals, stakeholders, and incentives. As a result, the capabilities they develop — and whether these are beneficial or harmful — will depend on tasks and contexts that may extend far beyond those anticipated during design, making it difficult to predict and govern the broader societal effects. As with other disruptive technologies, then, the full impact of these effects will only become visible once the systems are embedded in society. What is clear, however, is that this impact may be profound. Integrating highly autonomous systems into society could reshape how we work, care for others, and produce knowledge. These changes may be beneficial, but they also raise concerns about accountability for harm and over-reliance on systems in roles they cannot adequately perform. The central challenge, then, is how to develop such autonomous robots ethically despite being unable to fully anticipate their societal effects before deployment.

## 2.2 The Value Alignment Problem

The core capacity of autonomous systems — to decide and do things by themselves — forms at once their most promising as well as their most dangerous feature. While it promises that we can delegate more and more tasks to these systems, it, at the same time, exposes us to the risk that their execution of these tasks will not be to our liking or even puts us in harm's way.[2] To avoid this outcome, researchers have focused on the question of *how* we can ensure that the activities of autonomous systems and the goals that they pursue remain in line with human values — a question known as the *value alignment problem* (e.g., Gabriel 2020; Zhi-Xuan et al, 2025).[3]

[1] Setting final goals is a characteristic of human or full autonomy, which is not at issue here.

The strength of this approach is that rather than diminishing the autonomy of these systems, it *redirects* it. By teaching autonomous systems what we find important, or by providing them with a set of values that should be kept into account while acting, the idea is that even when we cannot predict what actions the system will take or what goals it will pursue, they will be aligned with our values and thus be acceptable. This way, or so the thought goes, we can preserve the benefits that we might derive from integrating autonomous assistants into our society, whilst neutralizing the risk that their actions will harm us.

While the idea behind this approach is promising, it rests on the condition that we are able to align the system's activities and decision-making with our values. And this condition, in turn, rests on us being able to either (1) develop systems that are capable of moral reasoning (i.e., *artificial moral agents*) or, (2) to develop a sufficiently broad set of values, including rules on how to prioritize them, that can guide autonomous systems in evaluating their best course of action in any situation they encounter. As we will argue in this section, both options face a number of challenges that make the current approach to value alignment, as a possible answer to how we can develop autonomous robots in an ethically sound way, unlikely to succeed.

**Option 1: Creating Artificial Moral Agents.** One proposed way to make sure that the activities of autonomous systems align with our values would be to create not just autonomous systems but *artificial moral agents*: systems whose behavior is governed by the same moral standards that govern our own behavior. We thereby use the term 'agent' in its philosophical understanding, so as referring to systems that are able to act in a manner that is both operationally (moving by oneself) and motivationally (self-movement on the basis of self-set goals, desires, intentions, etc.) autonomous (e.g., Anscombe 1957; Davidson 1963; Frankfurt 1978; Steward 2012).

A moral agent, now, is an agent whose actions are "governed by moral standards" (Himma 2009, p. 21). It can differentiate the morally right from the morally wrong, and act accordingly. Important thereby is that letting its actions be guided by moral standards is a *choice* of the agent itself. A moral agent is therefore a genuinely moral being. Having this capacity, we can assign it duties and obligations and hold it accountable when it neglects them. The idea is that in creating artificial moral agents, such systems would not just align with our values but actually *share* them.

The question is how feasible this idea really is. From the technical side, there is the question whether we can actually create moral agents. Himma (2009), for example, has argued that this would entail being able to create something akin to artificial consciousness. Given the fact that there is no universally agreed upon account of what consciousness is and that it is still debated whether current state-of-the-art systems even are agents (e.g., Prunkl 2022; Floridi 2025; van Lier 2025), the technical realization of artificial moral agents seems not very realistic.

Besides the challenge of technical realization, there are also ethical considerations that need to be taken into account, such as whether we really want to create a system that has agency or even something akin to consciousness, given our current instrumental use of artificial systems. And, even when we manage to create moral artificial agents, there is no guarantee that their values would be the same as ours, especially when these systems would be autonomous. It is therefore likely that the value alignment problem would persist, making this not a feasible option.

**Option 2: Developing a Moral Framework.** Given the technical and ethical challenges of creating systems that are genuinely moral in and of themselves, many scholars working on the value alignment problem have opted instead for the approach of teaching or encoding in autonomous systems the values that their activities should align with. This approach faces its own set of challenges, however. Drawing on Gabriel (2020) and Zhi-Xuan et al. (2025), four main challenges emerge. The *challenge of authority selection* concerns who may legitimately choose the framework when universal participation is impracticable. The *challenge of value selection* concerns how to choose between a narrow set of values that leaves considerable room for interpretation and a broader set that may contain conflicts. The *challenge of genuine value change* reflects the evolution of values over time. Finally, the *challenge of technical realization* concerns whether the full complexity of values, the related background assumptions, and possible exceptions can be adequately encoded in or taught to artificial systems.

---

[2] This is the case even when excluding the possibility, however remote, that autonomous systems might autonomously select their own objectives (or override those assigned to them), in the course of carrying out delegated tasks.

[3] Both Gabriel (2020) and Zhi-Xuan et al. (2025) make an excellent case as to why autonomous systems should align with our values rather than with our preferences. Given the scope of this paper, we direct the interested reader to their work and assume from here on out that the main aim of the value alignment problem is to align the actions of autonomous systems with our *values*. We thereby follow Gabriel and Zhi-Xuan et al. in defining a value as an evaluative concept or judgment that indicates what someone or something morally (vs. legally, rationally, etc.) 'ought to do' as defined by individuals and society.

These four challenges show that efforts to align autonomous systems with human values will remain imperfect because misalignment incidents cannot fully be prevented. The authority that selects the values might not represent all views (*the challenge of authority selection*), we might exclude values that are deemed important or prioritize them in a manner not accepted by all (*the challenge of value selection*), values might change over time without being accounted for (*the challenge of genuine value change*), or we might not be able to teach the moral framework in all its complexity to the system or encode it within it (*the challenge of technical realization*).

Thus, even if the system's learned behavior closely aligns with our values, it is both theoretically and practically impossible that it will be fully aligned. Moreover, maximizing alignment assumes that the closer a system is aligned, the more acceptable its behavior will be. However, even a very closely aligned system might still learn behavior that is unacceptable to us. The aim to maximize alignment is thus not sufficient for developing autonomous systems in an ethically sound way.

**A shift in perspective.** Given the aforementioned challenges, we now argue that to develop autonomous robots in an ethically sound way, a change of perspective is needed. This change of perspective entails acknowledging that full alignment is not possible and that misalignment incidents can therefore not be prevented in their entirety. In acknowledging this fact, the question becomes how we can ensure that, when such misalignment does occur, it is *acceptable* in the sense that it will be of minimal consequence to society.

This is a method that we are already familiar with. Imagine for example the construction of a playground. Given that such a playground is meant for children, it is of the utmost importance that it is *safe*, while still providing entertainment for them. However, no matter how one builds this playground, there is always the risk that a child might fall or hurt itself. After all, children are unpredictable and curious, and they have little sense of danger. What one can do, then, is to make sure that if a child falls, it cannot fall from great heights or fall on a hard underground. One can put warnings for parents to stay around, especially for the more complex constructions, and make sure that there are no sharp edges. While complete safety cannot be guaranteed, one can make sure that when something goes wrong this is a very rare occurrence of minimal consequence. What one aims for, then, is that the safety risks are *acceptable*.

We now claim that to develop autonomous robots in an ethically sound manner, a similar approach needs to be adopted: Rather than pursuing the maximization of alignment, we should instead aim to make the occurrence of misalignment incidents acceptable. In the next section, we develop this notion of *acceptable misalignment* in a more detailed way.

## 3. Acceptable Misalignment as Limited, Justified, and Modifiable Residual Risk

In this section, we argue that misalignment is acceptable when the *residual risks of harm* that remain, after alignment methods and safety measures have been implemented, are *limited* and *justified*, and the assessment of these risks remains *modifiable*. We base this main claim on three sub-claims: (1) Just because all the available alignment methods and safety measures have been taken, this does not mean that any remaining risk of potential misalignment should be accepted; (2) Only when the remaining potential for misalignment is limited and justified, and this assessment is modifiable, can it be deemed acceptable; and (3) Research should focus on developing and implementing alignment and safety mechanisms that are able to limit and justify residual risk, as well as make its assessment modifiable.

In Section 3.1, we address sub-claims (1) and (2) by introducing the concept of residual risk and defining the conditions under which residual risk is credibly limited and justified, and its assessment remains modifiable. In Section 3.2, then, we address sub-claim (3) by going deeper into what criteria such alignment and safety mechanisms need to meet.

### 3.1 Residual Risk After Risk Treatment.

Let's go back to the playground example. Even when a playground has been carefully designed, children may still fall. Yet it would be misleading to say that the playground is safe simply because most falls have limited consequences. We would also want to know how often children are likely to fall, whether some parts of the playground expose them to more serious injuries, who is most exposed, and whether further precautions could reasonably have been taken. The same applies to autonomous systems. The fact that an incident of misalignment is unlikely, or that its most immediate consequences appear limited, does not by itself make the system safe.

The notion of *residual risk* provides a more precise way of expressing this point. The NIST AI Risk Management Framework (RMF) defines residual risk as the risk that remains after risk treatment has been applied. It is the risk that is left after available measures have been taken to avoid, reduce, constrain, or otherwise manage this risk. The NIST AI RMF further stresses that residual risks affect not only the organization deploying a system, but also end users and affected communities. It therefore calls for these risks to be documented, communicated, monitored, and compared with an explicitly defined risk tolerance (National Institute of Standards and Technology, 2023). A similar direction is taken in Article 9(5) of the Artificial Intelligence Act (Regulation (EU) 2024/1689, 2004), which requires providers, within the legally mandated risk-management process, to ensure that the relevant residual risk associated with each hazard, as well as the overall residual risk of a high-risk AI system, is judged acceptable.

This judgment concerns the system as it is used in its intended context, under conditions of reasonably foreseeable misuse. Residual risk is therefore not the harm that remains after something has gone wrong, but *the potential for harm that remains after appropriate risk-treatment measures have been applied*. While these frameworks provide an important starting point, they do not define what 'acceptable' means. The NIST AI RMF deliberately leaves risk tolerance to organizations, communities, regulators, and sector-specific norms and standards. The AI Act also requires an assessment of acceptability without, however, fully determining the principles through which it should be reached.

This is where the present paper aims to contribute. Its purpose is not to introduce residual risk as a new concept, but to connect it with the alignment problem and with a staged process for governing autonomous systems. In our proposed view, misalignment is not acceptable simply because it is inevitable. Rather, we claim that it becomes acceptable only when the risks it produces have been reduced and contained as far as reasonably possible, when the decision to tolerate what remains can be publicly justified, and the assessment of these risks remains modifiable. In the following, we explain what we mean by this.

**The limitation of residual risk.** A residual risk is limited when *there are credible limits to the harm that the system can produce*. Likelihood and severity are central to this assessment, but they are not sufficient. The assessment must also consider the number and vulnerability of those exposed, the duration and reversibility of possible effects, the resources available to the system, and the possibility that several minor risks could accumulate to generate a serious social impact. A low probability does not make a catastrophic or irreversible harm acceptable, while a relatively frequent failure may be tolerable if its effects remain contained and are easily repairable.

**The justification of residual risk.** A risk, even when deemed low, may still be imposed unfairly: A system might expose only a limited number of people to harm, for example, while placing the entire burden on workers or vulnerable users. The distinction between social acceptance and ethical acceptability is crucial in this context. As Taebi (2017) argues, the fact that a technology is accepted by an organization, a market, or even a majority does not show that its risks are ethically acceptable. A technology may simply be accepted because affected groups lack information, or realistic alternatives. Whether or not a residual risk can be justified, therefore, does not only depend on how much risk remains, but also on *who bears it, who has the power to accept it, and who is able to obtain protection or redress*.

**Acceptability as a modifiable judgment.** It might be that with the introduction of autonomous robots in society changes occur whose effects we cannot fully foresee and that might impact whether or not the system can still function in a way that is acceptable to us. For example, with the rising temperatures due to climate change, incidents are now occurring in which children have burned themselves on playground equipment. With climate change, then, a new factor emerged that influences whether or not equipment can be deemed safe for children. To be able to take newly emerging factors into account, a final condition for acceptability is therefore that *its assessment remains modifiable over time*.

### 3.2 Requirements for Alignment and Safety Mechanisms

Having set the normative requirements for *acceptable misalignment* in Section 3.1, we are now faced with the question of what kind of alignment and safety mechanisms can meet these requirements for autonomous robots (sub-claim 3). The question is thus what methods would allow us to develop these systems in such a way that the potential occurrence of misalignment incidents are credibly limited and justified, and it remains possible to reevaluate this assessment.

To limit and justify residual risk, and retain the option to modify this assessment, we first need to be able to *identify* this risk. The question, however, is how one is able to identify this risk for a system whose core characteristic is that it can act *autonomously*, meaning that what it will end up doing is always to an extent *unpredictable*. As we saw in Section 2.2, current approaches to the alignment problem try to solve this difficulty by attempting to make these systems act *like us*, in the sense that their actions are governed by the same values that govern our own. The problem with this approach is that we put the burden of determining what the right action is on the system *itself,* and this requires a level of technical sophistication that will be difficult, if not impossible, to realize. But what if we do not have to put this burden on the system itself?

Drawing on relationships between humans and animals, Darling (2021) argues that the question of how to deal with "autonomous agents that cause unanticipated harm based on their own decision making" is not a new one (p. 80). Animals are autonomous agents that can neither understand nor act in a way that fully aligns with our values, yet we have managed to not just live with them but even to partner up with some of them and treat them as our companions. According to van Lier (2025), this is possible because animals are agents that *act unpredictably in a predictable manner*. While we cannot fully anticipate what they might do because of their autonomy, animals are still bound by what is possible for them to do given their embodiment, what they have done and learned before, and what kind of action-potentials are afforded to them by their environment. Considering these factors enables us, to a certain extent, to *anticipate* what animals can and, importantly, *cannot* do, and to adapt both our own behaviour and the environments in which we interact with them to limit harmful encounters. We thus do not expect animals to act like us, but instead rely on our own ability to assess what kind of interactions with them are possible and under what conditions.

We now propose that we should take a similar approach in dealing with autonomous robots. It is thereby important to note that we are not saying here that autonomous systems are agents, nor that we must aim to build (moral) agents (see also Section 2.2). Rather, we argue that we should make use of our own ability to identify what possible harms might occur in dealing with autonomous systems and to adjust our interactions with them accordingly. This way, misalignment is not caused by the system 'misbehaving' but by things we can more readily control for such as technical malfunction, known uncertainties, human behavior, or triggers in the environment. To make this possible, however, it is necessary that we can anticipate, to an extent, what these systems can and cannot do. A first criterion for alignment and safety mechanisms, therefore, is that *they make the unpredictable functioning of the system predictable*.

Still, even when we might be able to anticipate, to an extent, the capabilities and functioning of an autonomous system, we cannot fully anticipate how it will be received in society. For example, the system could learn certain things or be used in a way that changes how we do things. And this, in turn, might lead to new risks that cannot be foreseen at design time. To ensure that these systems are developed in an ethical manner, it is therefore important that the possibility exists to deal with newly emerging risks.

Here we turn to van de Poel (2016), who argues that the introduction of socially disruptive technologies into society should be treated as a social experiment. Such experiments, like testing a new medicine or policy, entail many unknowns, because they involve structural social changes, the consequences of which cannot be fully foreseen. To prevent catastrophic consequences, social experiments are, ideally, conducted in a step-by-step well-documented process of controlled trial-and-error, where human well-being takes center stage and the option exists to adapt or stop when residual risks cannot be credibly limited or justified.

We now argue that residual risks can be identified, assessed, and dealt with by adopting a similar approach for the introduction of autonomous systems into society. An additional benefit of this approach is that it enables us to spot early on when residual risk changes, and adapt our treatment of it on the basis of this assessment. A second criterion for alignment and safety mechanisms is therefore that *they make it possible to treat the introduction of autonomous systems into society as a social experiment*.

In the next section, we present a new framework, called the *Acceptable Misalignment Framework*, which consists of several alignment mechanisms that, together, are able to meet the aforementioned two criteria.

## 4. The Acceptable Misalignment Framework

In this section, we introduce the *Acceptable Misalignment Framework* (Figure 1), a step-by-step workflow of alignment processes aimed at achieving the main goal of reducing misalignment to an acceptable level, meaning that *the residual risk that remains is credibly limited and justified, and the assessment of this risk can still be modified*.

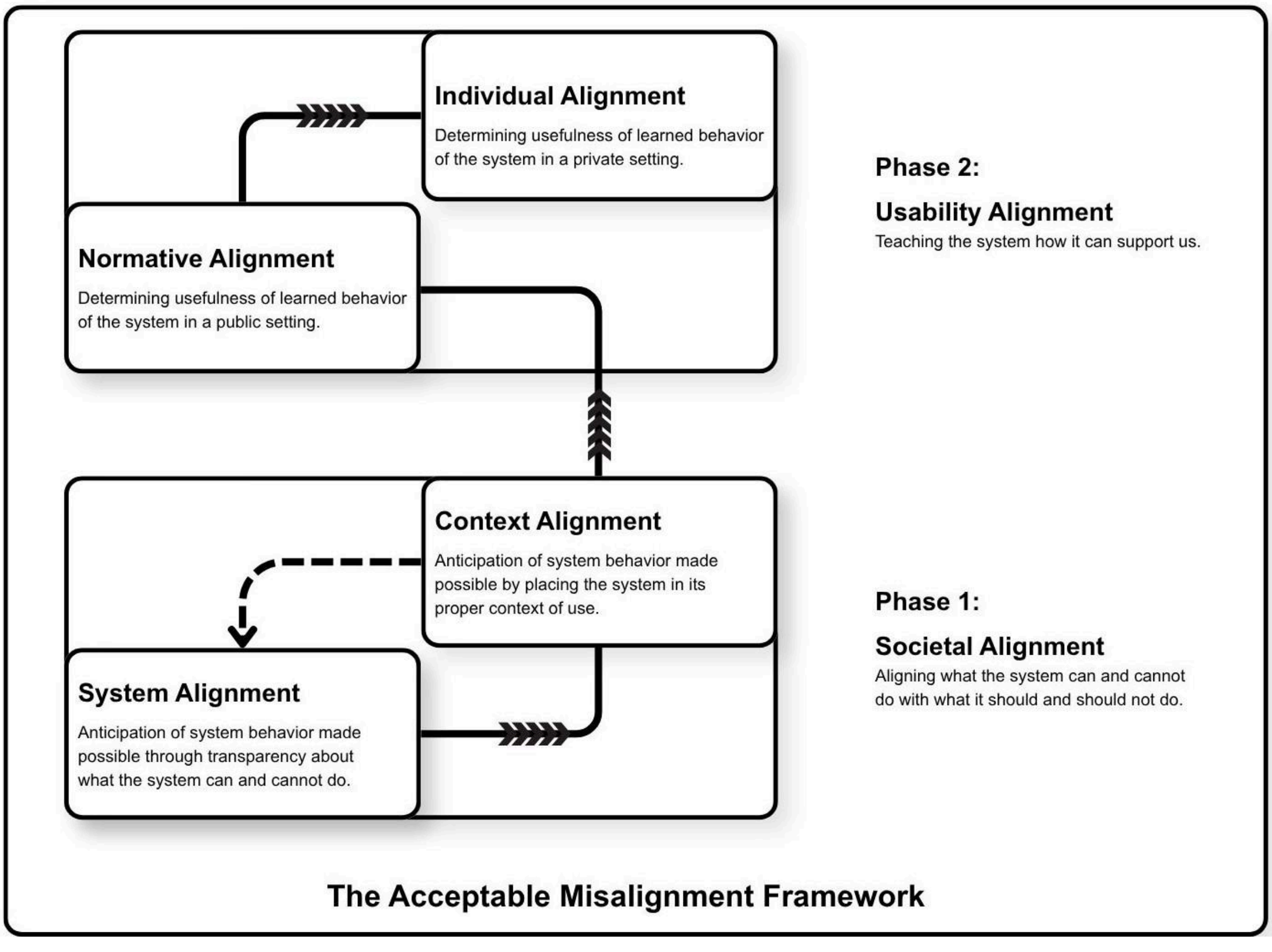

**Fig. 1** The *Acceptable Misalignment Framework* represents a step-by-step workflow (represented by the multi-headed arrows) consisting of multiple alignment processes executed in two phases which are aimed at bringing the risk of misalignment to an acceptable level. Before an autonomous robot should be allowed in public and private settings, it must be possible, to an extent, to anticipate what the system is likely to do on the basis of its form and the environment that it is in. If this is not yet possible in the *context alignment* stage, the system's functioning has to be further calibrated and tested by returning to the *system alignment* stage (dotted arrow).

The Acceptable Misalignment Framework consists of two phases, each further divided into two additional stages. During the first phase, which we call *Societal alignment*, the focus is on developing (*System alignment*) and socially embedding (*Context alignment*) autonomous robots in such a way that it becomes possible for us to anticipate, to a certain extent, *the manner in which* they might function unpredictably (Criterion 1, Section 3.2). While we might not be able to predict fully what these systems might do given their ability to act autonomously, we can, minimally, make transparent what the systems can and cannot do in a more general sense, as well as what uncertainties about their functioning remain. This way, it becomes possible to interact with these systems safely in the sense that we can identify and deal with potential risks by either adjusting our own behavior or by putting safety precautions in place in the context in which the system is to be deployed.

Making the system safe for use is not sufficient, though. Additionally, we want autonomous systems to be *useful*. What makes them useful, however, is something that we need to *discover*, as it depends on what is technically possible in the context in which the system is used and whether or not unforeseen issues arise once the system is deployed in real-life situations. In the second phase then, which we call *Usability alignment*, the focus is on how we can, over time, align the system's learned behavior with the way that we want the system to support us. Since this process might create potentially dangerous situations, this phase is again divided up in two stages, where the system's learned behavior is first evaluated in public settings (*Normative alignment*) before it is permitted for use in private settings (*Individual alignment*).

In approaching acceptable misalignment in this manner, the framework presents alignment as a step-by-step incremental process, linking particular forms of alignment to the various stages of the societal integration of autonomous systems. During each step, the system is aligned as far as reasonably possible, given what is required for that stage of the integration process, and the residual risk of potential harm is limited and justified. Given that this is a gradual process, the residual risk can be (re)assessed at each step and the approach can be stopped or modified when necessary. This procedure thereby anticipates that new problems might pop up at later stages which cannot be foreseen during the design stage. Effectively, then, the Acceptable Misalignment Framework represents a workflow for a *social experiment* (Criterion 2, Section 3.2), through which autonomous robots can be introduced into society in such a way that the residual risks that remain, after all available alignment and safety measures have been taken, are acceptable to society.

In the next sections, we discuss the framework in more detail.

## 4.1 Societal Alignment

The aim of societal alignment is to align what the system *socially* should and should not do with what it *technically* can and cannot do. The objective is thereby to place these systems in a context in which their particular skill-set can be put to good use and no misconceptions can arise about what the system can and cannot do. In this phase, then, *safe* Human-AI interaction takes central stage, where we hold that interactions are safe when it becomes possible for humans to *anticipate* what such systems are likely to do — when we can predict, to an extent, the *manner* in which autonomous systems (might) behave unpredictably — and, on this basis, to adjust either our own behavior or to adapt the context in which the system is to be deployed in such a way that human intervention remains effective and the system can be shut down or returned to a safe state.

To make this possible, we argue that two processes of alignment are necessary, one in which the goal is to make the autonomous functioning of the system anticipatory (*System alignment*), and a consecutive one in which the goal is to match the system with its appropriate context(s) of use (*Context alignment*). In the following, we describe the way that we envision these processes in more detail.

### 4.1.1 System Alignment

Autonomous robots are able to interact with the physical or digital environment in which they are placed through a *medium* (see Section 2.1). Examples of such mediums are wheels to drive around, a gripper to pick and place things, an interface or avatar, or a voice or facial expressions to communicate. The chosen medium thereby provides autonomous systems with a particular *repertoire* of actions that they can use to interact with their environment and the entities within it. These mediums can, and often do, serve a second function, however. In choosing a particular medium or labeling the medium in a certain way, the designer can signal to us what repertoire is available to the system, what it can and cannot do, and what kind of interactions we can accordingly expect to have with these systems. A humanoid robot, for instance, can be expected to sit, walk, gesture, or even communicate with us when it is given facial expressions and a voice. We refer to this signal as the system's *form*.

The chosen form indicates not only *what kind* of actions the system is capable of, but also *how* it is able to execute these actions. So we expect a humanoid robot to be able to pick up a box (since we can) and that it will pick up this box in a similar manner as we would have. Depending on the medium that is chosen, then, certain expectations are created not only about what the system can do but also about how the system is able to do things. Problems can now arise when there is a mismatch between what the form suggests the system to be capable of and what the medium actually allows the system to do. An example is the Atlas Humanoid Robot of Boston Dynamics. The form of this robot is human-like. Its medium, however, is human-like *plus*: It has more degrees of freedom than we humans have. This allows the robot to stand up in ways that will never be possible for us humans. While this is very impressive, potentially harmful situations can occur because based on its form, we cannot fully anticipate the entire range of behaviors that this system is capable of, leading us to either under- or overestimating what the system technically can and cannot do.

We now propose that the risk of the occurrence of this form of misalignment can be limited through what we call *System alignment*, which consists of the following actions:

1. **Technical alignment:** To align the system's technical capacities with its intended function by making it robust[4] and, where possible, fault tolerant,[5] and to provide evidence of this through testing.
2. **Form alignment:** To align the system's form with its medium, and vice versa. This can entail the following:
    a. **Aligning the system's form with its medium:** Removing features that might create ambivalent expectations about what the system is actually capable of (e.g., eyes that cannot see, a human voice, labeling the system as a 'friend').
    b. **Aligning the system's medium with its form:** Constraining, altering, or expanding the system's repertoire to fit the desired form in a way that resembles the entity on which the form is based.
    c. **When a mismatch cannot be avoided:** To provide demos and explicit descriptions that show how the chosen form should be interpreted and to what extent it diverges from the medium.

During the system alignment stage, then, the residual risk of the system malfunctioning is limited by making the technical capacities robust and where possible fault tolerant. As for the residual risk of harms occurring because of the under- or overestimation of the system's technical capacities, this is *limited* by aligning the system's form as far as reasonably possible with its technical capacities, both by fitting the form to the technical capacities, and vice versa. The residual risk that remains in both cases is *acceptable* when it is made transparent and can be *justified* by disclosing known and/or suspected epistemic gaps in the developer's ability to make the system robust or to estimate users' expectations and their particular causes, as well as by acting on these epistemic gaps through additional testing, narrower operating conditions, stronger safeguards, or non-deployment when the residual risk cannot be credibly limited.

#### 4.1.2 Context Alignment

In the system alignment stage, the focus was on reducing misalignment incidents to an acceptable level that related to the system itself. However, focusing only on the system itself is not sufficient. This is because even when these incidents are brought to an acceptable level, further misalignment incidents might occur when the system is placed in a context of use that it is not suited for, given its technical capacities. Here we can think of a chat-bot that is asked for medical advice or a human driver in a hurry who ignores a traffic sign, causing a self-driving car to wrongly predict the driver's behavior, with a traffic incident as a result. In these cases, the system is either assigned a (social) role by its user that does not align with its technical capacities or the environment itself contains unforeseen elements that prevent the system from performing its service in a correct manner.

[4] A system's technical capacities are robust when it can deliver a correct service (i.e., operation, output or task) through its medium despite adverse situations, particularly in uncertain environments (such as an unexpected obstacle on a robot's path).
[5] A system's technical capacities are fault tolerant when it can deliver a correct service despite faults affecting system resources (e.g., sensor malfunction or software fault)

We refer to the (physical or digital) environment in which the system is active and the (social) role that has been assigned within it as the system's *context of use*. We thereby define the environment as the physical and/or digital space that the system can affect through its medium and the role as the intended (social) function the system is assigned in this environment. The aim of *Context alignment*, now, is to make the residual risk that derives from a mismatch between the system's technical capacities and its context of use acceptable. It is based on the idea that once we know what a system can and cannot do, and what epistemic gaps about its particular functioning persist (which were established during the system alignment stage), we are able to determine what, accordingly, the system *should and should not do* in a given context of use, including the conditions under which potential misalignment between the system's capacities and its context of use is still acceptable.

Determining the conditions under which a system can be used in a particular context requires the expertise of those who have knowledge about *the social embedding* within which the system executes its assigned role — those who can identify the relevant stakeholders and their needs in a particular context of use and that know *how* the role should be executed to meet these needs. Most of all, it requires the expertise of those who can recognize challenges that cannot be solved technically, but rather require organizational, social, or governance solutions.

Context alignment is therefore not a one-person job: It requires the shared insights of a group of experts that have specialized knowledge of the social, institutional, legal, ethical, and operational environment in which the autonomous robot is to be deployed. Here one can think of domain specialists, practitioners, regulators, social scientists, ethicists, and other stakeholders. Only such a group of experts can determine, first, what is actually required for a system to perform its assigned role well within a particular context of use, and second, the conditions under which potential misalignment incidents between the system's functioning and these requirements are still acceptable.

To perform this job, the group first has to identify possible cases of misalignment, given their expert knowledge about the context of use in question. Here the importance of the system alignment stage comes to the fore, as it should form the basis for the identification of potential misalignment incidents. It might for example be that given the system's form, misconceptions could arise about what role it is supposed to play, or that the context's environment is far more complex than the settings in which the system's operations were tested. Furthermore, it might be that there exist epistemic gaps about the reliability of the system's safety mechanisms that cannot credibly be justified, especially when the system might interact with vulnerable people or handle sensitive information. The expert group is thus tasked with listing — based on the documentation delivered by the developers in the previous system alignment stage — all the possible misalignment incidents that might occur when the system is deployed in a particular context of use.[6]

This list is also intended to allow for the assessment of the likelihood and severity of the occurrence of each of the potential context misalignment incidents. Based on this evaluation, the experts can either decide that the residual risk is (1) acceptable, in the sense that it can be credibly limited and justified, and its assessment modified over time; that it is (2) unacceptable and that the system should therefore not be deployed in the context of use; or (3) that the residual risk *could* become acceptable depending on whether the residual risk is reduced to an acceptable level.

Whether or not the group deems the residual risk acceptable depends on a variety of reasons, such as the perceived value of the system's function for society, the difficulty of identifying misconceptions and/or preventing them from happening, and the complexity of determining the impact of the use of a particular system on society. If the system's function is deemed to be valuable to the context of use, but misconceptions might arise about its assigned function, the expert group can decide that the environment should be 'reshaped' (enveloped) according to the limitations that the system has in executing its intended role.[7] In practice, this process can involve either the creation of a digital or physical space in which the system can be tested securely — *a sandbox* — or the regulation of the conditions under which interaction with the system is allowed.

[6] Naturally, potential incidents will be missed. The goal is, however, that by combining the technical knowledge of the developers with the knowledge of these experts, a relatively exhaustive list can be drawn-up.

[7] Naturally, this requires insight in what the system can and cannot do, so that it is possible to anticipate possible triggers or obstacles in the system's environment. This is why system alignment should precede context alignment.

It should be noted, however, that enveloping the environment should only be pursued when the system is the preferred option for performing the assigned role and the performance of this role has value for the context of use. When this is not the case, for example when the system is less qualified to do the job than humans, envelopment should be avoided as it might be detrimental to the environment in question because of transfer costs, additional labor, accessibility burdens, or restrictions that are placed onto workers. If transferring the function to the system could still be of benefit to society (for instance when there are labor shortages), then the better option would be to require that the system *itself* is further adapted to meet the standards for functioning well in that context of use. This means a return to the system alignment stage until the provided standards are met.

As shown in Figure 1, the *Societal alignment* phase thereby becomes an iterative process. Even when, based on what the system can do, it has been determined that it should not be used in a particular context, this does not mean that future use in that context might not become possible at a later point of time. It might for example be that stakeholders in a particular environment become with time better attuned to what the system is capable of (technical literacy), or that the system can be developed in a more secure way. Within this evolving landscape, Context alignment thus does not just involve making a decision about whether a system fits a particular context of use. It also sets standards for alignment processes and oversees whether these standards are met. In doing so, Context alignment serves three functions: (1) To make the residual risk linked to context misalignment acceptable; (2) To strengthen the environmental and contextual safeguards designed to protect users in the event of system failure; and (3) To provide clear baselines for regulation on the promotion and use of autonomous systems.

## 4.2 Usability Alignment

Autonomous robots can support humans in social and versatile contexts such as retail, homes, hospitals, museums, and schools. However, to properly support humans in this way, it is not sufficient that these systems do not harm us or that we can anticipate in a reliable manner what they are likely to do. What we need additionally is that these systems are actually *useful* — that they can learn *how* we want them to support us.

Naturally, this is not immediately clear. We might have a certain idea of how a system can support us, but this might not be practically realizable either because of the system's technical capacities or because the environment does not afford the system with the option to do so. It might also be that the learned supportive behavior is only helpful to some but inconveniences, or even harms, others, and is therefore not acceptable overall. Or, it could turn out that what we imagined we needed is actually not that helpful, but that the system can assist us in other ways. The 'usefulness' of a system is thus something that in part needs to be *discovered*, something that is learned *over time*, and that thus necessarily needs to be subjected to a process of trial-and-error. The question is how this process can be controlled in such a way that the residual risk of (1) something going wrong during this discovery process or (2) of the system being useful to some but not to others is acceptable.

This question is placed at the center of the second phase of the Acceptable Misalignment Framework, which we refer to as *Usability alignment*. During this phase, the aim is for the system to become useful to us by learning, through a number of iterative stages, how we want it to support us. We now argue that to reduce the residual risk of something going wrong during the discovery process (1), the process of trial-and-error first needs to take place in an *organizational setting*, where the possibility exists to gradually broaden the context in which the system is active, to put up safeguards at each stage, to use sandboxes, to formulate clear norms by linking them to the organizational setting, and to oversee and subject the learning process to multiple overseers.

We further argue that the learned behaviors that were phased out during this stage, which we call the *Normative alignment* stage, should form the normative boundaries for an authorized envelope of permissible behaviors that the system can learn in individual support settings (during the *Individual alignment* stage). This way the risk is reduced that the system learns behavior that is only useful to the individual user but that is inconvenient or harmful to others (2).

While this trial-and-error process might reduce the residual risk of something going wrong during the learning process and of the system being useful to some but not to others, this does not yet mean that it is acceptable. Additionally, the risk should be *justified* in that the learning phase took place in the two aforementioned stages (1) and by providing in the first stage credible reasoning as to why it is acceptable that the system might be useful to some but not to others (2). Finally, the assessment of the acceptability of this residual risk should remain modifiable by retaining the oversight in the organizational setting and by making it possible to update the authorized envelope. Only under these conditions, we argue, can *usability misalignment* be deemed acceptable.

In the following, we go deeper into what each of the two final alignment stages entail.

#### 4.2.1 Normative Alignment

During the *Normative alignment* stage, the aim is for an autonomous robot to become useful to a multi-stakeholder group linked to a particular organizational setting by aligning its learned behavior with the values, constraints, responsibilities and standards that this group has set for its behavior. With *organizational setting*, we thereby refer to an institutionally organized environment in which people work together to pursue a common mission through coordinated roles, established practices, and shared resources. Examples of such organizational settings include museums, research institutes, schools, industrial enterprises, hospitals, and government organizations.

During the normative goal alignment stage, an autonomous robot should be introduced into an organizational setting in a gradual manner. Which setting is selected will thereby depend on the appropriate context(s) of use of the system, given its technical capacities (which was established during the societal alignment phase). The advantage of first introducing the robot in an organizational setting is that: (1) This setting allows for a clear identification of the relevant stakeholders; (2) The values, constraints, responsibilities and standards that this group sets for the robot's behavior are linked to the organizational setting; and (3) The organizational structure provides a bounded deployment environment which allows for a gradual expansion of the area or domain in which the system can be active, with multiple levels of human oversight and intervention options.

Within this organizational setting, then, it becomes possible to determine what learned behavior of the robot is useful and therefore permissible, and what behavior is not useful, inconvenient, or even potentially harmful and should therefore be phased out through a process of controlled trial-and-error. Important thereby is that the robot is introduced in a gradual manner in the organizational setting, so that at each level of integration it can be clearly defined what role the system is supposed to fulfill and according to which norms, which stakeholders should set these norms, what oversight mechanisms are implemented, and who is responsible for this oversight. This way, it becomes possible to treat the introduction of the robot in the organizational setting as a *social experiment*, where the well-being of the stakeholders takes center stage and where at each level of integration it remains possible to phase out unwanted behavior, adapt the integration method, or even stop it when issues occur.

The aim of this social experiment is to end up with *an authorized envelope of permissible behaviors* that the robot can learn. It is this envelope that sets the boundaries for what a robot can learn once it is introduced in a private setting. The objective thereby is to establish what learned behaviors are acceptable to all and which are unacceptable in the sense that they might only benefit some while inconveniencing, or possibly harming, others. Once this authorized envelope has been drawn up, and binds the learning behavior of the robot, we posit that it could be safe enough to be introduced in private settings.

#### 4.2.2 Individual Alignment

Only those systems that have been attuned to the collective judgments of developers, experts and stakeholders during the previous alignment stages should be made available for private use and then only within the context of use that they have been assigned to. If the previous alignment processes have gone well, sufficient safety guardrails should have been drawn up to ensure that the system will not be able to learn behaviors that put the individual user at risk.

Naturally, not all risks can be avoided. It might for example be that the system malfunctions due to a fabrication error or some other unexpected technical malfunction, or that the individual user manages to use the system for a function it was not meant to perform. Here, though, we can take inspiration from the way that we handle biological agents. Over the years, we have allowed certain animal species to be kept as pets. A distinction is thereby made between pets that everyone is able to keep (hamsters, certain breeds of dogs, cats, etc.) and those for which their owners need a special license and/or training. When a dog is aggressive or an exotic pet is held without a proper license, we have agencies in place to deal with such cases, for example by euthanizing the pet, removing it from its owner, or by revoking the owner's license to keep such pets.

We now argue that for the individual use of autonomous robots, similar rules and regulations can be drawn up. As a result, when the system happens to malfunction, a team monitoring these cases can neutralize the system and remove it from the home. Additionally, insurance schemes can be drawn up that provide compensation in cases where harm occurs due to such malfunction. As for cases in which individuals use a system outside of the scope it was meant for or misuse specialized autonomous robots for which training or a license is required, the particular individual can be held accountable and punished accordingly, for example by removing their license or putting them on a blacklist.

Given the aforementioned alignment processes that proceed individual use, however, these cases should be outliers rather than the norm. What is more, since these alignment processes require extensive documentation about the particular uses of the system, there should be a substantial basis for policy makers to identify such potential outliers and draw up suitable rules and regulations to deal with them. Taken together, then, the alignment processes that are part of the Acceptable Misalignment Framework should ensure that, when followed, misalignment cases will be acceptable in the sense that they are of minimal consequence to society.

## 5. Conclusion: General Outlook, Limitations and Uses

Socially disruptive technologies are technologies which may seriously impact society, either positively or negatively. What these impacts might be, however, is largely unknown and very hard to predict beforehand. We argued that autonomous robots capable of open-ended learning form such technologies. This is because we cannot fully foresee or test for all the things these systems might be able to learn in real-life situations, nor can we know what kind of situations they will encounter or how these might affect what these robots end up doing. Just as with socially disruptive technologies, then, we face the challenge of how we can develop these systems in an ethically sound way when the exact nature of their impact on society will only become known to us once they are introduced into it.

In this paper, we questioned the efficacy of current approaches that focus on maximizing the alignment of systems with human values, as these are unable to account for whether, and why, the risk of misalignment that remains after available alignment mechanisms have been implemented would be acceptable to society. We propose instead a new perspective, which shifts away from the maximization of alignment towards the conditions under which misalignment becomes acceptable. We explored what this shift entails by providing a definition of acceptability, drawing up the normative requirements that alignment mechanisms need to meet in order to satisfy our standards of acceptability, and by proposing a new methodological framework that is able to meet these criteria, which we call the *Acceptable Misalignment Framework*.

While we think that this framework is implementable, a lot of work remains to be done. One question, for example, is how these alignment strategies can concretely be implemented in current societal practices. Who will bear responsibility for overseeing the execution of the various stages and what kind of legislative measures can be taken to ensure that these stages are followed? And how can this framework be useful for both research purposes as well as commercial initiatives?

There are also more general questions about the scope of the framework. Where it is meant to provide concrete tools to deal with the unknowns of introducing autonomous robots into society, we have not looked at autonomous systems more generally. What would change once we use this framework for autonomous systems that are not, or less tangibly embodied, such as LLM-based systems? And can this framework more broadly be implemented for other socially disruptive technologies that are not necessarily autonomous? These are questions that are relevant and that should be addressed in future work.

However, we believe that the proposed framework and the concept of acceptable misalignment already provide fruitful input for the current debate on how to develop autonomous robots in an ethically sound way, paving the way for a perspective in which we treat these systems for what they are: Advanced tools capable of participating in social activities, albeit in their own limited, non-human way.

## Acknowledgements

This paper is the result of work carried out within the project PILLAR Robots, 101070381, HORIZON-CL4-2021-DIGITAL-EMERGING-01.